%% file: main.tex
\documentclass[letterpaper, 10pt, journal]{IEEEtran}
\usepackage{amsmath,amsfonts}
\usepackage{algorithm}
\usepackage{array}
\usepackage[caption=false,font=normalsize,labelfont=sf,textfont=sf]{subfig}
\usepackage{textcomp}
\usepackage{stfloats}
\usepackage{url}
\usepackage{amsmath}
\usepackage{verbatim}
\usepackage{graphicx}
\usepackage{cite}
\usepackage{booktabs}
\usepackage{hyperref}
\usepackage{tikz}
\usetikzlibrary{shapes.geometric, arrows.meta, positioning, calc}
\usepackage{diagbox}
\usepackage{makecell}

\title{Task Parameter Extrapolation via Learning Inverse Tasks from Forward Demonstrations}

\newcommand{\serdaradd}[1]{\textcolor{black}{#1}}
\newcommand{\serdardel}[1]{}

\newcommand{\finaldiff}[1]{\textcolor{black}{#1}}

\begin{document}

\markboth{IEEE Robotics and Automation Letters. Preprint Version. Accepted June, 2026}
{Bahar \MakeLowercase{\textit{et al.}}: ShortTitle}

\author{Serdar Bahar$^{1}$, Fatih Dogangun$^{1}$, Matteo Saveriano$^{2}$, Yukie Nagai$^{3}$, Emre Ugur$^{1}$%
\thanks{Manuscript received: March, 6, 2026; Revised June, 3, 2026; Accepted June, 27, 2026. This paper was recommended for publication by Editor T. Ogata upon evaluation of the Associate Editor and Reviewers’ comments. This work was supported by the EU through projects \textit{INVERSE} (no. 101136067) and \textit{HARMONICA} (no. 101294331), and in part by the World Premier International Research Center Initiative (WPI), MEXT, Japan.} 
\thanks{$^{1} $Department of Computer Engineering, Bogazici University, Istanbul, Türkiye. $^{2} $Department of Industrial Engineering, University of Trento, Trento, Italy. $^{3} $IRCN, the University of Tokyo, Tokyo, Japan. Corresponding author: {\tt\footnotesize serdarbahar44@gmail.com}}%
\thanks{Code is available at \url{https://github.com/serdarbahar/task_par_extrapolation}}
\thanks{Digital Object Identifier (DOI): see top of this page.}
}

\maketitle

\begin{abstract}
Generalizing skill policies to novel conditions remains a key challenge in robot learning. Imitation learning methods, while data-efficient, are largely confined to the training region and consistently fail on input data outside it, leading to unpredictable policy failures. Alternatively, transfer learning approaches offer methods for trajectory generation robust to both changes in environment and tasks, but they remain data-hungry and lack accuracy in zero-shot generalization. We address these challenges in the context of task inversion learning and propose a novel joint learning approach to achieve accurate and efficient knowledge transfer. Our method constructs a common representation of the forward and inverse tasks, and leverages auxiliary forward demonstrations from novel configurations to successfully execute the corresponding inverse tasks, without any direct supervision. We demonstrate the extrapolation capabilities of our framework through ablation studies and experiments in simulated and real-world environments that require complex manipulation skills with a diverse set of objects and tools, where we outperform diffusion-based \serdaradd{and multimodal VAE alternatives.}


\end{abstract}


\input{introduction}

\input{method}
\input{experiments}
\input{real_world_experiments}

\input{limitations}
\input{conclusion}

\bibliographystyle{IEEEtran}
\bibliography{ref}

\end{document}

%% file: introduction.tex
\section{Introduction}

A fundamental challenge in robotics is developing methods for efficient, generalizable skill acquisition. An autonomous robot is expected to adapt to variations in its environment and changes in task or skill specifications. This requires skill policies that can generalize beyond training conditions, which remains a significant obstacle for current learning paradigms.

Data-driven approaches in robotics have demonstrated impressive generalization capabilities, but often at the cost of extensive data collection and interaction with the 
environment~\cite{Jang2022}. 
Transfer learning frameworks aim to mitigate this by leveraging knowledge across tasks. A common technique, domain randomization, focuses on creating policies that are robust to variations in the environment, such as changes in lighting, friction, or object 
textures~\cite{OpenAI2020}. 
However, these approaches are mostly designed to handle environmental changes within a single task definition. While other methods, such as cross-domain transfer, exist, they often require additional training in the target domain and struggle with data efficiency and robust 
zero-shot performance~\cite{Niu2024}.

Imitation learning (IL) offers a more data-efficient alternative by leveraging expert demonstrations for the task of interest. 
Deep learning has given rise to powerful and flexible methods such as Conditional Neural Movement Primitives ~\cite{CNMP} and Stable Movement Primitives~\cite{StableMP}, which learn complex, non-linear skills directly from demonstration data. More recently, advancements in deep generative modeling have enabled robotic policies to represent rich distributions using diffusion- and flow-based models~\cite{chi2024diffusionpolicy, zhang2024flowpolicy, nguyen2025flowmp}. Although these methods are  effective, their operational strength lies in \textit{interpolation}, allowing them to accurately generate behaviors similar to the expert demonstrations provided, but they consistently fail at \textit{extrapolation}~\cite{Ross2011}. This inability to generalize beyond the training data results in unpredictable trajectories for novel inputs, often compromising the success of downstream tasks that rely on the trained policy. While some recent work has shown promise in learning explicit mathematical relationships to enable extrapolation~\cite{Villeda2023}, the problem remains largely unsolved for mainstream IL frameworks.


We explore the paradigm of joint learning as a mechanism for zero-shot extrapolation, specifically within the structured domain of \textbf{task inversion}. Many robotic skills naturally exist as forward-inverse pairs.\serdardel{(e.g., pushing an object to a goal and pulling it back; assembling a mechanical component and disassembling it for repair).} \serdaradd{In many such pairs, a forward demonstration is more readily available than its inverse counterpart (e.g., a component is often observed being assembled before it is disassembled), motivating the acquisition of inverse skills without direct inverse supervision.} We observe that given a common representation of forward and inverse tasks, an inverse task for a novel environment configuration can be generalized from a provided forward task for a similar configuration (e.g., learning to disassemble a component by observing its assembly). Following this idea, we propose a novel joint learning framework that effectively transfers knowledge from a forward task to its inverse counterpart by learning a common representation for both. Specifically, excluding out-of-distribution generalization for action generation, we focus on inferring inverse executions whose sensorimotor (SM) patterns are similar to an observed set of SM demonstrations. For instance, a robot that acquired pushing and pulling skills for a set of objects, can utilize our framework to infer how to pull a novel object after observing how the object is pushed. 


To achieve this, our method extends the framework of Conditional Neural Processes (CNP)~\cite{CNP} and Deep Modality Blending Networks (DMBN)~\cite{DMBN}. CNP~\cite{CNP} is a meta-learning network that models complex data distributions through a Gaussian process-based perspective and has proven instrumental for skill learning in the robotic domain~\cite{CNMP, ACNMP, DMBN}. In particular, DMBN~\cite{DMBN} is a multi-modal architecture that jointly learns different modalities of robot actions with CNP by forming a shared latent space of multi-modal observations. Our contributions include:
\begin{enumerate}
    \item A joint learning framework that enables zero-shot extrapolation to novel task parameters for an \textit{inverse task} by leveraging auxiliary demonstrations from its corresponding \textit{forward task}.
    \item A training methodology, including a demonstration matching algorithm based on initial and final environment states and an interleaved training schedule, allowing a flexible data demonstration and efficient learning from datasets including auxiliary demonstrations.
    \item Separating task parameter conditioning from SM encoding, crucial for generalizing to unseen parameters.
    \item \serdaradd{An extensive evaluation across synthetic, simulation, and real-world settings; demonstrating the necessity of forward-inverse pairing, robustness to observation noise, and superior extrapolation performance over diffusion-based and multimodal VAE baselines.}
\end{enumerate}

%% file: method.tex
\section{Method}

We are given a demonstration set that includes forward executions of skills and their inverse counterparts, along with their task parameters, where forward and inverse demonstrations are not necessarily matched, and not all counterparts are observed. Inverse executions represent executions that induce the opposite state transition in the environment. Our aim is to infer and generate inverse executions for novel task parameters by leveraging auxiliary forward demonstrations available for those parameters. The executions may or may not be direct reversals of the forward SM trajectories, depending on the complexity of the task. \serdaradd{Here, we briefly introduce the background, then outline our problem formulation, framework overview, and skill training and generalization methods.}
\subsection{Background}
  
\subsubsection{Conditional Neural Processes (CNP)}

Given a set of input data, CNP\cite{CNP} estimates a Gaussian process conditioned on a varying number of input observations. To achieve this, CNP employs a shared parameter encoder network, $E_{\phi}$, to encode observations into a fixed-size representation, $r$, and a decoder network, $Q_{\theta}$, which produces Gaussian distributions conditioned on $r$ with input $x_q$. CNP computation can be expressed as $
\mu_q, \sigma_q = Q_{\theta}(x_q \oplus \frac{\sum^{n}_{i=1}E_{\phi}(x_i, y_i)}{n})$, where $\mu_q$ and $\sigma_q$ are the mean and standard deviation of the predicted Gaussian distribution, respectively, $\oplus$ is the concatenation operator, and $\{(x_i, y_i)\}_{i=1}^{n}$ are input observations to the network. 


\subsubsection{Deep Modality Blending Networks (DMBN)}
DMBN\cite{DMBN} was proposed as a neural action recognition mechanism, where the robot's motor patterns were automatically triggered by observing the corresponding actions executed by itself or other robots. The robot jointly learns an action across different modalities, such as joints and images, and employs multiple encoder and decoder networks, each accounting for a different modality. The training and inference of DMBN are similar to those of CNP, except that an aggregation operation across modalities is used. 
DMBN can be conditioned on any desired image at any time step, and can generate a motor trajectory consistent with the desired conditioning in parallel, avoiding the accumulation of prediction errors. 


\begin{figure}[tb!]
    \centering
    \includegraphics[width=0.75\columnwidth]{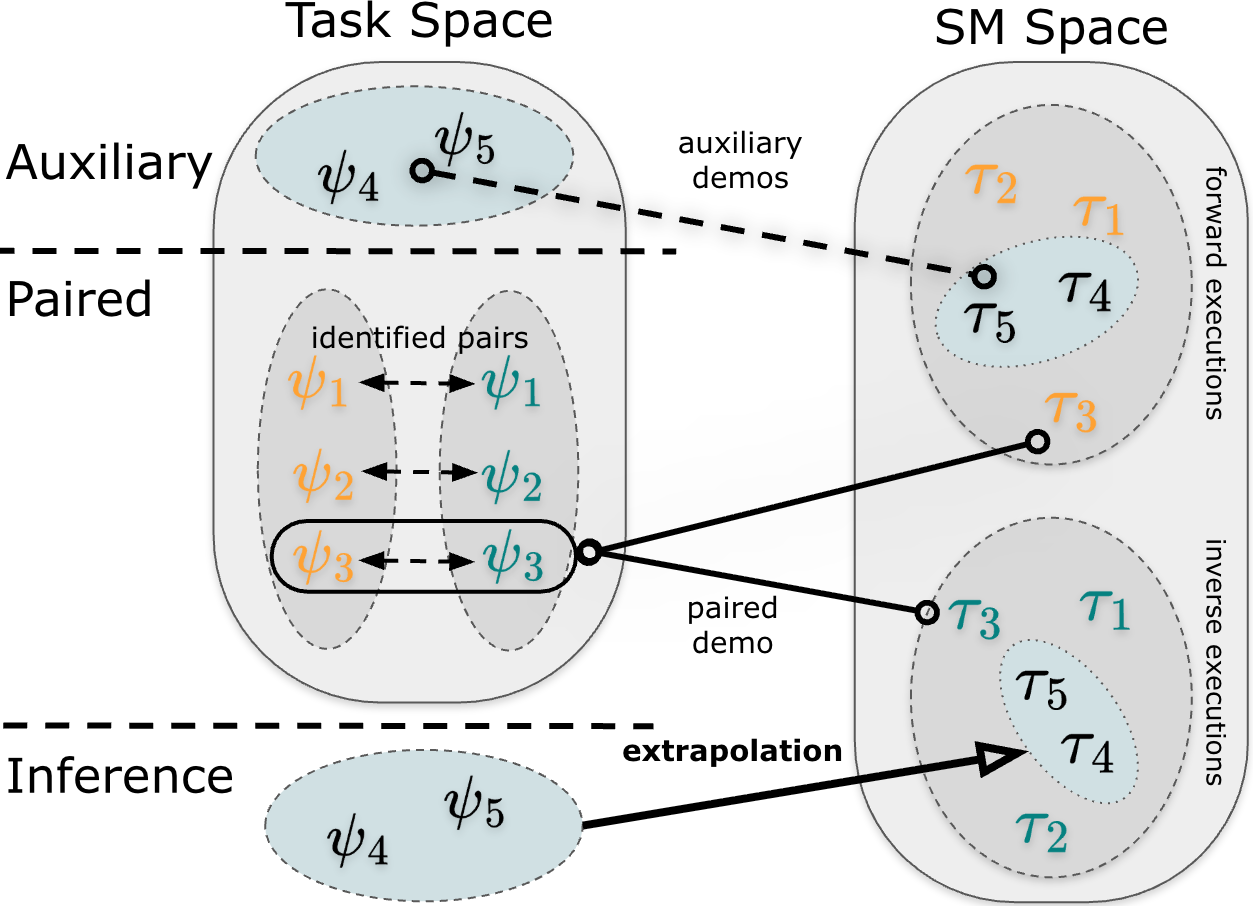}
    \caption{ The demonstrations are decoupled into task and SM spaces. In the top, demonstrations are visualized as lines between task parameters ($\psi$) and SM executions ($\tau$). Our method will first identify correspondence between demonstrations and learns a common representation of the forward and inverse tasks using paired and auxiliary datasets. The learning objective is to infer inverse executions for task parameters drawn from the auxiliary distribution.
    }
    \label{fig:problem_definition}
\end{figure}



\subsection{Problem Definition}

Formally, we define a forward task as one that causes a state transition from $S_{init}$ to $S_{final}$, and its corresponding inverse task as one that reverses this transition, from $S_{final}$ to $S_{init}$. The demonstrations are defined by a tuple that includes $S_{init}$ and $S_{final}$, the SM trajectory ($\tau$) observed during this transition, and the corresponding task parameter ($\psi$), which might be a scalar value or an image, depending on the task. The initial unorganized sets of forward and inverse demonstrations are denoted by $\mathcal{D}_F = \{(\tau_i^F, \psi_i^F, S_i^{F,init}, S_i^{F,final})\}_{i=1}^{N}$ and $\mathcal{D}_I = \{(\tau_i^I, \psi_i^I, S_i^{I,init}, S_i^{I,final})\}_{i=1}^{N}$, respectively. Furthermore, the dataset for the forward task is augmented with an auxiliary set of $M$ additional demonstrations, denoted as $\mathcal{D}_{aux}^F = \{(\tau_j^F, \psi_j^F, S_j^{F,init}, S_j^{F,final})\}_{j=N+1}^{N+M}$. \finaldiff{As illustrated in Fig.~\ref{fig:problem_definition}, the objective is to infer inverse task executions for task parameters drawn from the same distribution as those in $\mathcal{D}_{aux}^F$.} Our system is trained with forward-inverse pairs of task parameters and SM trajectories (numbered 1-3 in the figure), and forward-only OOD task parameters and in-distribution SM trajectories (numbered 4-5 in the upper part of the figure). We address the task extrapolation problem by inferring the SM trajectories of inverse tasks for OOD task parameters, i.e., novel task parameters that were not provided in the inverse demonstrations. As shown on the right (blue ellipses), we expect the queried task parameters to be in-distribution with the auxiliary task parameters, and the SM trajectories to be within the distribution of the demonstrated SM space. 


\subsection{Method Overview}

Our method proceeds in two stages. In the first stage, a paired demonstration dataset is constructed by identifying forward and inverse demonstrations that are counterparts. Such pairings are crucial, as the common representation relies on explicit correspondences to capture the relationships between task parameters and their forward and inverse executions. The resulting dataset of paired forward and inverse demonstrations ($\mathcal{D}_{paired}$) is illustrated with dashed lines in Fig.~\ref{fig:problem_definition}. In the second stage, skill training operates on this paired dataset and the auxiliary forward dataset. The training requires (1) separate encoding of the task parameter to avoid failures on novel task parameters at inference time, and (2) joint training over both the paired and auxiliary demonstrations to form a common latent representation while enriching it with auxiliary information. With these two conditions, the trained model can infer inverse executions for task parameters drawn from the auxiliary distribution, as illustrated at the bottom of Fig.~\ref{fig:problem_definition}.

\subsection{Identifying Forward-Inverse Pairs}
\label{subsec:identify_pair}

\finaldiff{Establishing forward-inverse execution correspondences (paired dataset, $\mathcal{D}_{paired}$) in the unorganized sets of forward and inverse demonstrations ($\mathcal{D}_F$ and $\mathcal{D}_I$) can be formulated as a linear sum assignment problem.} Given the two initial demonstration sets, $\mathcal{D}_F$ and $\mathcal{D}_I$, we construct a cost matrix $C \in \mathbb{R}^{N \times N}$ where each entry $C_{ij}$ quantifies the dissimilarity between the final environment state of the i-th forward demonstration and the initial environment state of the j-th inverse demonstration. The optimal bijective pairing is then obtained by minimizing the total assignment cost.\serdardel{In our experiments, where the environment is represented as state-based vectors, we use Euclidean distance as the dissimilarity measure. Scenarios involving higher-dimensional environment representations, such as images, would require a more specific similarity metric.} \serdaradd{In our experiments, as similarity metrics, we use Euclidean distance for position-based features and structural similarity index measure (SSIM)~\cite{ssim} for images.} Note that the auxiliary set $\mathcal{D}_{aux}$ is excluded from this process. The problem can be solved using the Hungarian algorithm~\cite{kuhn}, and the resulting bijection is used to create a paired dataset for joint learning. 

\begin{figure}[tb!]
    \centering
    \includegraphics[width=0.85\columnwidth]{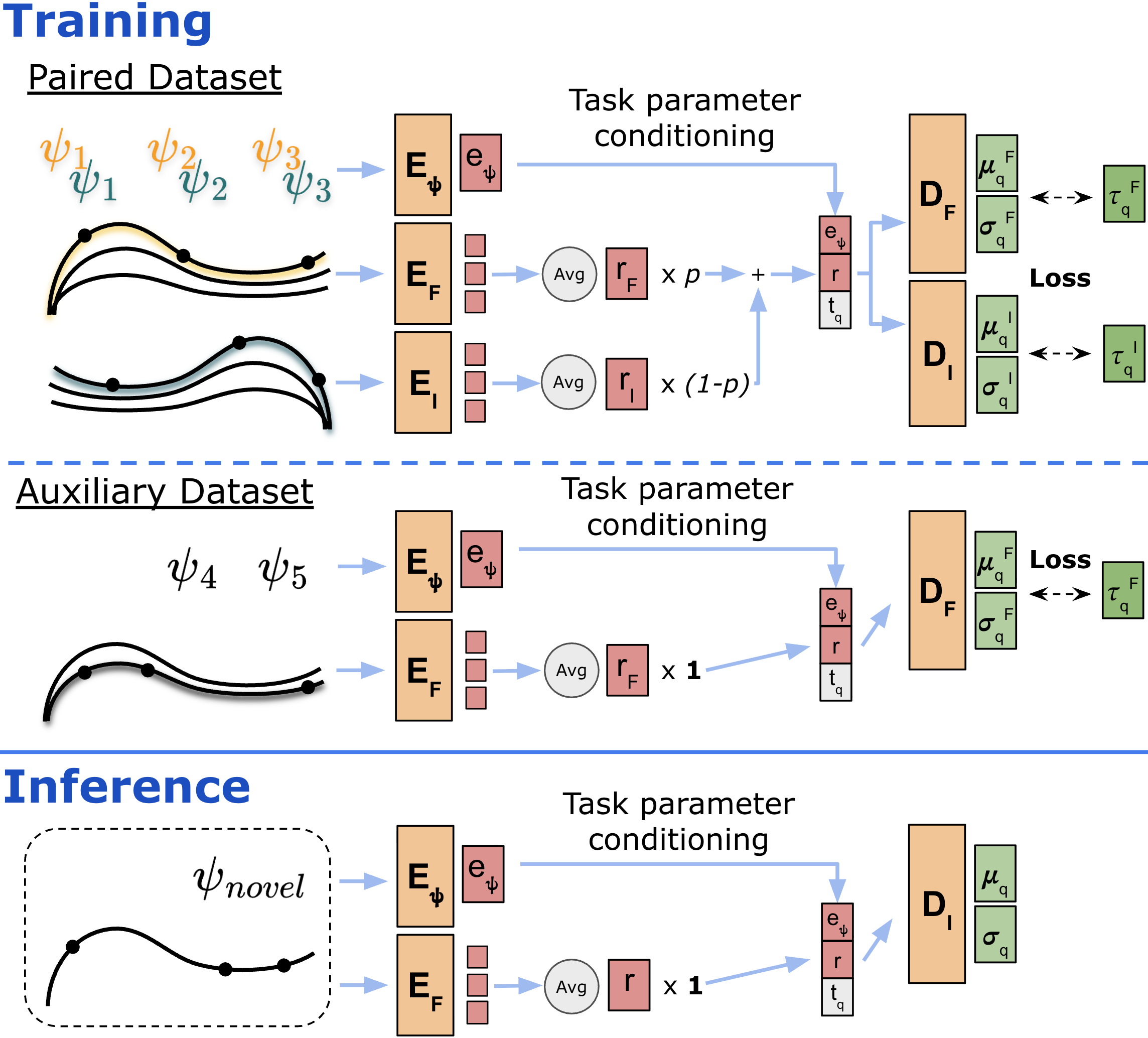}
    \caption{
        Overview of the proposed method. The first and second rows show the overall paired and auxiliary passes, respectively, during training. The third row shows the inference conditioned on the novel task parameter and observations from a forward execution to generate the full inverse execution.}
    \label{fig:proposed_method}
\end{figure}

\subsection{Skill Training for Task Parameter Extrapolation}

{\bf Training with paired} forward-inverse trajectories is illustrated in the upper row of Fig.~\ref{fig:proposed_method}. For training, first, a pair of forward-inverse demonstrations ($\tau^F$ and $\tau^I$) is randomly chosen from the dataset. Then, random observation points (empirically set to range from 1 to 15) are selected from the paired forward and inverse demonstrations. These observations (shown with 3 dots in the figure) are independently sampled with time-steps normalized to a $[0, 1]$ time scale. Next, these observations are \serdaradd{encoded} by separate forward and inverse encoders ($E_F$ and $E_I$), and outputs are averaged to yield representations for the pair ($r_F$ and $r_I$). These representations are aggregated into a unified representation ($r$) via a convex combination with a stochastic weight: $r = p \times r_F + (1-p) \times r_I$, where $0 \le p \le 1$, as shown in the figure. A separate module ($E_\psi$) is employed to embed the task parameter ($\psi$), where the embedding ($e_{\psi}$) is generated by a fully connected and a convolutional neural network for vector and image inputs, respectively. This embedding ($e_{\psi}$) is concatenated with the common latent representation ($r$) and a randomly chosen query time step, and then passed through both forward and inverse decoders ($D_F$ and $D_I$) to predict SM values for the corresponding time step, as shown on the right side of the upper row in the figure. The decoders output a normal distribution's mean and variance, optimized via negative log-likelihood against ground truth. This repeats with random forward-inverse pairs and observations until convergence. After training, we set $p=1$ and query all time points to generate the full inverse SM trajectory from a chosen forward trajectory.

{\bf Training with auxiliary} forward trajectories without inverse counterparts is shown in the second row of Fig.~\ref{fig:proposed_method}. We incorporate auxiliary data via interleaved training with two forward-pass types. In the \textbf{paired pass}, trajectory pairs are sampled from $\mathcal{D}_{paired}$, and their common latent representation is formed by a random convex combination. In the \textbf{auxiliary pass}, samples from $\mathcal{D}_{aux}$ form the common latent representation solely from forward-task SM observations, with $p=1$, while $E_I$ and $D_I$ remain frozen. This integrates OOD task parameters into the common latent space without changing the representation capacities of $E_F$ and $D_F$, since these parameters are expected to produce in-distribution SM executions.
The two passes alternate stochastically with probability $p_{\text{aux}}$, balancing SM-execution learning from paired task parameters with integration of novel task parameters and their unpaired forward executions. $p_{\text{aux}}$ was set empirically.


{\bf Inference of inverse} trajectories for novel parameters is shown in the bottom row of Fig.~\ref{fig:proposed_method}. Here, our aim is to generate the trajectory of a novel inverse task by querying our system with the novel task parameter and observation points from its forward execution. As shown in the figure, these observations are \serdaradd{encoded} by $E_F$ and then averaged to generate the common representation ($r$), which can be used to generate both forward and inverse SM trajectories \serdaradd{using $D_F$ and $D_I$, respectively}. Next, $r$ is used as input to $D_I$, along with the task parameters, to generate the complete inverse SM trajectory by querying the system at all time steps from 0 to 1. 

%% file: experiments.tex

\section{Experiments}

\serdaradd{We evaluated our method through
experiments with synthetic, simulations, and real-robot data, and compared it with Diffusion Policy (DP)~\cite{chi2024diffusionpolicy}-inspired and Multimodal VAE (MVAE)~\cite{sejnova2024mvae} joint learning alternatives. All reported significance values were obtained using paired t-tests.}


\subsection{Synthetic Data}

\serdaradd{We designed a synthetic scenario to analyze the importance of identifying forward-inverse pairs, and our method's sensitivity to two main hyperparameters: auxiliary pass probability and maximum observation count.} \serdaradd{We assessed the {\it correspondence quality} on all paired datasets and {\it hyperparameter sensitivity} on perfectly paired datasets with auxiliary trajectories.} We model SM executions as explicit expressions, their inverses as time-reversals, and derive task parameters and environment states from these expressions.

\subsubsection{Experiment Setup}
The datasets consist of trajectories that are sinusoidal curves defined over a normalized time period $[0,1]$. For the forward task, we define the trajectory as $\tau_F(t) = \psi \cdot \sin(\frac{3}{2}\pi t)+t$, and we use its time reversal, $\tau_I(t) = \tau_F(1-t)$, representing the inverse task, where task parameter ($\psi$) is the scalar amplitude ranging between 0.1 and 0.25. Also, for a trajectory $\tau$, the environment states ($S_{init}$ and $S_{final}$) are $\tau(0)$ and $\tau(1)$, respectively.

\begin{figure}[tb]
    \centering
    \includegraphics[width=0.85\columnwidth]{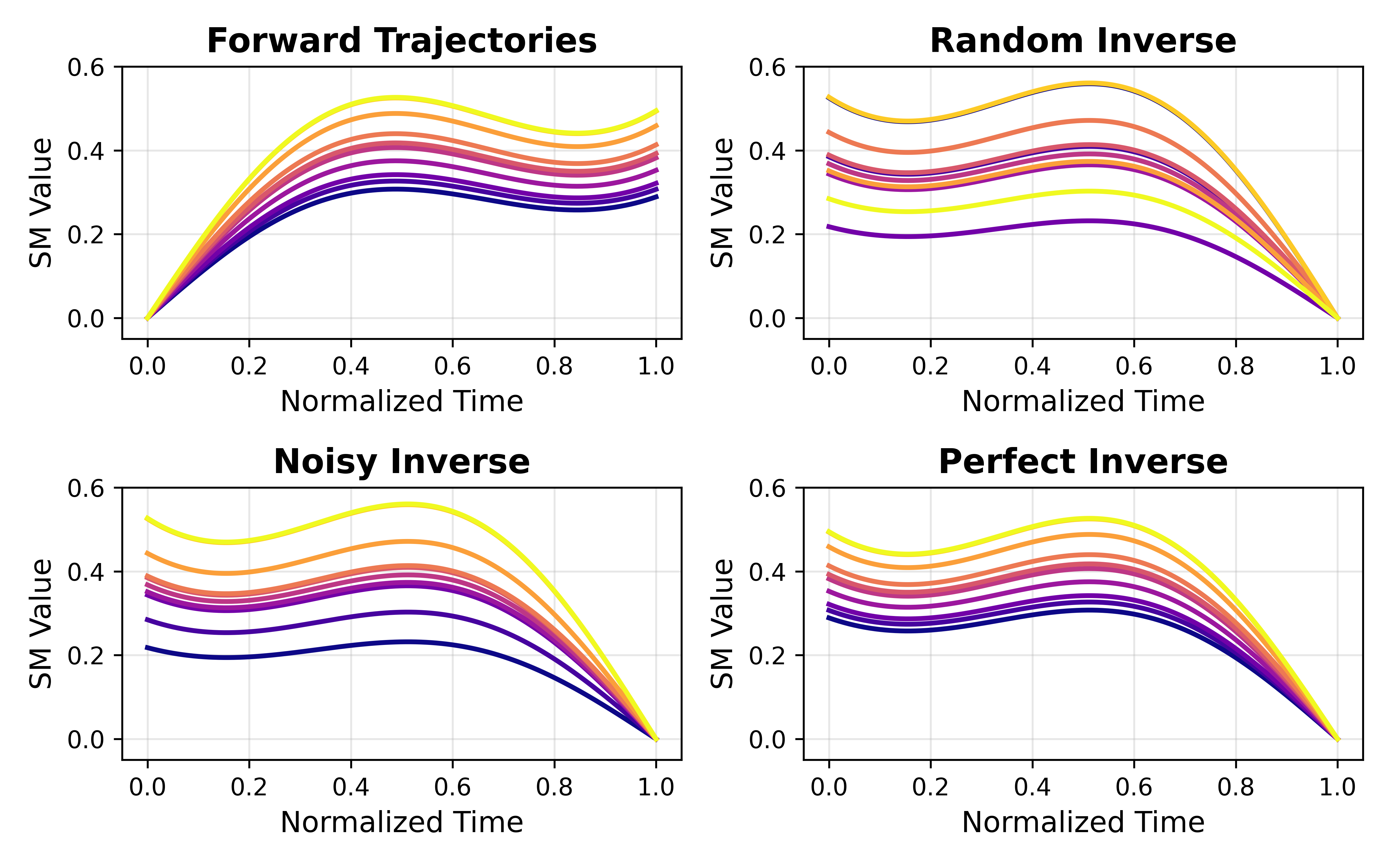}
    \caption{Synthetic datasets. The color correspondence illustrates the pairing.}
    \label{fig:syn_datasets}
\end{figure}

\paragraph{Datasets}
We designed four conditions that differ in how the demonstration pairs are matched, namely random, paired noisy, paired perfect, and uniform, and for each condition, we generated five datasets of \serdaradd{10 demonstration pairs each} (visualized in Fig.~\ref{fig:syn_datasets}): \textbf{Random:} Demonstrations in $\mathcal{D}_F$ and $\mathcal{D}_I$ were generated from randomly sampled amplitudes, and randomly paired. \textbf{Paired Noisy:} Demonstrations in $\mathcal{D}_F$ and $\mathcal{D}_I$ were generated from randomly sampled amplitudes, and paired using our proposed algorithm. \textbf{Paired Perfect:} Demonstrations in $\mathcal{D}_F$ were generated as in the previous conditions. Inverse task trajectories with the \textit{same} amplitudes were used to form $\mathcal{D}_I$. \textbf{Uniform:} Demonstrations in $\mathcal{D}_F$ were generated from uniformly spaced amplitudes, and their perfectly corresponding trajectories were used to create $\mathcal{D}_I$, mimicking ideally collected and paired demonstrations.
\serdaradd{Note that for the {\it sensitivity analysis}, five additional forward \textbf{auxiliary} datasets ($\mathcal{D}_{aux}$)  are generated using randomly sampled amplitudes from the same range. The auxiliary sets are combined with the \textbf{Paired Perfect} set, and a model is trained for each seed.}

\paragraph{Model}

\serdaradd{We trained a 170K-parameter MLP encoder-decoder for 60k steps with a batch size of 4 using the AdamW optimizer (lr: $5\cdot10^{-4}$, weight decay: $1\cdot10^{-3}$). For the sensitivity analysis, we performed a grid search over the auxiliary pass probability (0.1 to 0.9 in 0.1 increments) and the maximum observation count (1, 15, 50, 100). For the paired datasets, we fixed the maximum observation count at 15.}

\subsubsection{Results}


\serdaradd{To assess the models, we use 20 novel forward trajectories to generate corresponding inverse trajectories, and we compute the Root Mean Squared Error (RMSE) between the generated and ground-truth inverse task trajectories. In the paired datasets,} the model trained on the \textbf{Random} dataset performed the worst with an \serdaradd{RMSE of $7.23 \pm 6.05$ (in units of $10^{-2}$)}, while our matching algorithm \textbf{(Paired Noisy)} yielded a statistically significant improvement ($p<0.001$), reducing the \serdaradd{RMSE by over 60\% to $2.85 \pm 2.02$ (in units of $10^{-2}$)} and demonstrating that a correct correspondence is critical for learning the shared structure and generating trajectories. The models trained on the idealized datasets, \textbf{Paired Perfect} and \textbf{Uniform}, achieved significantly lower errors (approximately \serdaradd{5 and 11 times lower}, respectively) than in the \textbf{Paired Noisy} condition ($p<0.001$). These results suggest that structured demonstration pairing is not only beneficial but also a fundamental requirement for joint learning of tasks. Besides pairing, ideally collected and paired demonstrations also help the model to generate more accurate trajectories. 

\begin{figure}[tb]
    \centering
    \includegraphics[width=0.85\columnwidth]{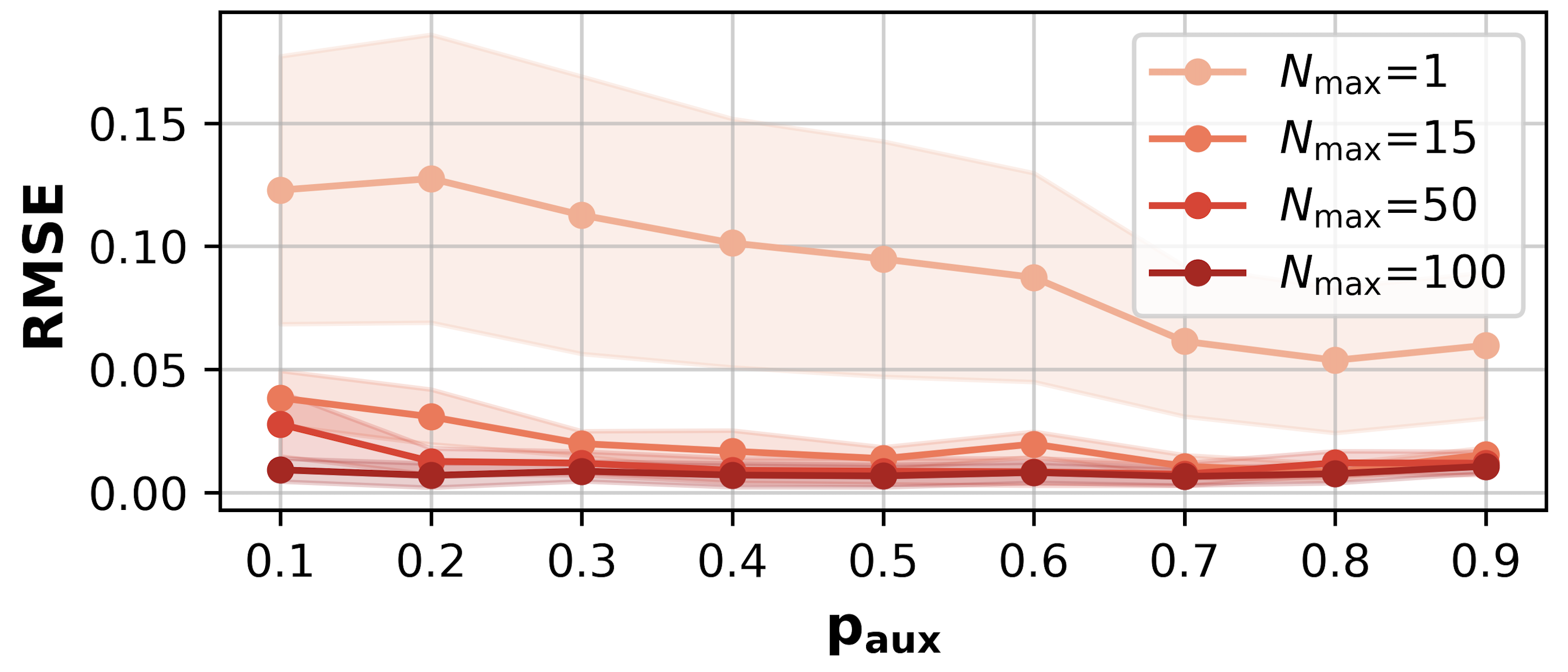}
    \caption{\serdaradd{Trajectory RMSE (mean $\pm$ std across five seeds) for models evaluated on the synthetic dataset with auxiliary trajectories, shown across different auxiliary pass probabilities ($p_{aux}$) and maximum observation counts ($N_{max}$).}}
    \label{fig:syn_sensitivity}
\end{figure}

\serdaradd{The sensitivity analysis of our method to the auxiliary pass probability ($p_{aux}$) and the maximum observation count ($N_{max}$) is illustrated in Fig.~\ref{fig:syn_sensitivity}. Increasing $N_{max}$ from 1 to 15 reduces mean RMSE by roughly two orders of magnitude (from $1.1 \times10^{-2}$ to $5\times10^{-4}$), while further increases to 50 and 100 yield minor gains at higher training cost. We therefore fix $N_{max}$ to 15. For different $p_{aux}$ values at observation count 15, mean RMSE varies from $4\times10^{-2}$ to $1.6\times10^{-3}$ ($25\times$), indicating a non-trivial effect. Specifically, the optimum $p_{aux}$ is in the 0.70-0.80 range, which is well above the proportion of auxiliary demonstrations in the dataset (10/30 $\approx$ 0.33). As a result, we empirically set $p_{aux}$ value via grid search also in the simulation and real-robot experiments.}

\subsection{Object Extrapolation in Robot Simulation}

We assessed the model's ability to extrapolate learned manipulation skills to entirely novel objects in a simulation environment, given paired demonstrations for two object categories (horizontal and vertical cylinders) and forward-only demonstrations for the test object categories (spheres and boxes). 
We also examined the role of the matching algorithm in both perfect and noisy data-collection scenarios, \serdaradd{its robustness under different noise conditions, and compared the proposed method against DP-inspired and MVAE alternatives.}

\input{algorithm_tikz}

\subsubsection{Experiment Setup}

\begin{figure}[t]
    \centering
    \includegraphics[width=0.85\columnwidth]{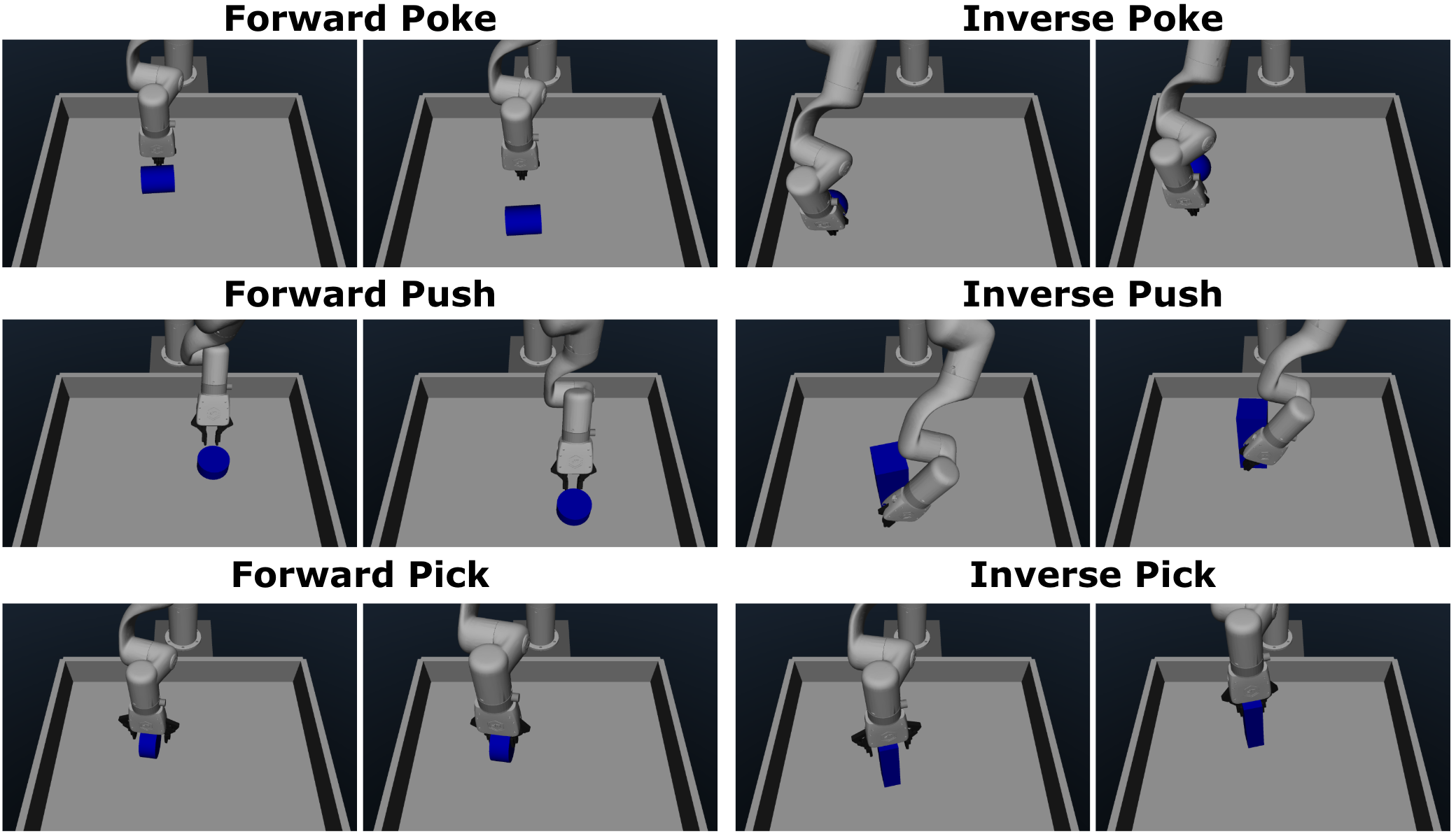}
    \caption{Example executions for the forward task (moving an object to a target) and their inverse counterparts with three different modalities for object categories (left to right, top to bottom: large horizontal cylinder, sphere, vertical cylinder, large box, small horizontal cylinder, small box).}
    \label{fig:sim_tasks}
\end{figure}

The experimental environment consists of a 7-DoF \textit{xArm 7} robotic arm with a gripper simulated in the MuJoCo~\cite{mujoco} physics engine. We define two related object manipulation tasks on a tabletop: \textbf{forward task}, manipulating an object from a designated start area to an end area of the table, and \textbf{inverse task}, manipulating the object in the reverse direction, which require distinct manipulation strategies due to the asymmetry in the robot's workspace. Visualization of demonstrations for the designated forward-inverse tasks are shown in Fig.~\ref{fig:sim_tasks}. The required robot action is determined by the object's physical properties, following the logic outlined in Fig.~\ref{fig:controller_logic}. The set of objects comprises vertical cylinders, horizontal cylinders, spheres, and boxes of varying sizes and orientations. The task parameter ($\psi$) is defined by a top-down depth image of the object and its initial position on the table. Lastly, the initial and final states ($S_{init}$ and $S_{final}$) include the object depth image and the position of the object.

\paragraph{Datasets}
To investigate the robustness of the pairing algorithm, we created the following datasets using joint angle measurements of the simulated robot: \textbf{Perfect Initial ($\mathcal{D}_F^{\text{Perfect}}, \mathcal{D}_I^{\text{Perfect}}$):} 60 demonstration pairs (20 for each action type) where all objects are \textbf{cylinders} and the task parameters for the forward and the inverse demonstrations are identical, leading to synchronized final and initial states for forward-inverse demonstrations, representing a perfect data collection scenario. \textbf{Noisy Initial ($\mathcal{D}_F^{\text{Noisy}}, \mathcal{D}_I^{\text{Noisy}}$):} 60 demonstration pairs with \textbf{identical objects}, Y-positions for demonstration for the tasks sampled independently, representing a noisy data collection scenario. \textbf{Auxiliary ($\mathcal{D}_{aux}$):} 20 demonstrations of the forward task, including randomly generated 4 spheres, 8 large boxes, and 8 small boxes with randomly sampled Y-positions.
\serdaradd{To further analyze the robustness of the pairing algorithm, we modified $\mathcal{D}_F^{\text{Perfect}}$ and $\mathcal{D}_I^{\text{Perfect}}$ with \textbf{observation noise}. We added three levels of Gaussian noise to observed object positions ($\sigma_{\text{position}} \in [0.25, 0.5, 1.0]$cm) and to images ($\sigma_{\text{camera}} \in [0.5, 0.75, 1.0]$cm), generating a total of 9 additional datasets with observation noise.}

\paragraph{Compared Models}
\label{compared_models}
We compared the proposed method \serdaradd{with DP-inspired and MVAE approaches in the Paired Perfect case.} The compared models are listed below:


\begin{itemize} 
     \item \textbf{Ours:} Proposed method with MLP encoders-decoders.
      \item \textbf{DP-Dual:} DP-based alternative to the proposed method. Two CNN-based decoders predicting forward and inverse joint angles with shared, global conditioning on the state information. Actions represented as joint angles in a single timestep, state information represented as action timestep, and task parameter.
     \item \textbf{DP-Mode:} Single CNN-based decoder. Actions represented as either forward or inverse joint angles in a single timestep, state information represented as action timestep, task parameter, and modality bit (forward or inverse). Note that the pairing of forward and inverse demonstrations is irrelevant to this model.
     \item \serdaradd{\textbf{MVAE:} Transformer-based encoders and decoders for the forward and inverse modalities, and MLP-based encoders and decoders for the task parameter modality. The model processes inputs to reconstruct the full forward and inverse trajectories, and the task parameter.}
     \item \serdaradd{\textbf{MVAE-Mask:} Extension of \textbf{MVAE}, joint angles at each time step are independently and randomly masked, forcing reconstruction of full forward and inverse trajectories from partial trajectory information, similar to ours.}
 \end{itemize}

The task parameter dimensionality is reduced to 75 by extracting features from a pre-trained ResNet50~\cite{he2016deep} and applying Principal Component Analysis. For statistical validity, we generated \serdardel{five}\serdaradd{eight} unique versions of Perfect and Noisy Initial datasets and used the same Auxiliary dataset along with them. We created four training conditions, including \serdaradd{the data-collection scenario (Perfect/Noisy) using the pairing strategy: Paired-Perfect (PP) and Paired-Noisy (PN) use our matching algorithm, while Unpaired-Perfect (UP) and Unpaired-Noisy (UN) use random pairing.} \serdaradd{To further test the efficacy of our matching algorithm, using three \textbf{Perfect Initial} datasets, we generated nine new variants for each, by modifying them with observation noise.} The models for the proposed method (940K trainable weights) are trained with $p_{\text{aux}} = 0.20$, batch size 24, for 80K steps using AdamW optimizer (lr: $5\cdot10^{-4}$, weight decay: $1\cdot10^{-3}$). The baselines DP-Dual (135M parameters), and DP-Mode (67M parameters) are trained with horizon 64, 8 predicted actions, batch size 256, for 200K steps with AdamW optimizer (lr: $5\cdot10^{-4}$, weight decay: $1\cdot10^{-6}$). DP-based baselines were developed using the LeRobot library~\cite{lerobot}. \serdaradd{These hyperparameters were selected through a grid search over model size, horizon, and number of predicted actions, with the chosen configuration already far larger than our model and consistent with the original Diffusion Policy~\cite{chi2024diffusionpolicy}. The baselines MVAE and MVAE-Mask (1.0M parameters each) are trained with batch size 16, latent dimension 32 for 500 epochs with AdamW optimizer (lr: $5\cdot10^{-4}$, weight decay: $1\cdot10^{-3}$). Masking probability for MVAE-Mask is set to $0.95$, yielding $10$ expected observations for each trajectory.}

\subsubsection{Results}
We evaluated on a test set of 10 spheres, 10 large boxes, and 10 small boxes, which were entirely novel and absent from training and auxiliary data. Inverse task trajectories are generated according to the determined task parameters, and their performance was measured by task success rate, trajectory, and final object position error based on the ground truth inverse task trajectory of the controller logic described in Fig.~\ref{fig:controller_logic}.

\paragraph{Effect of Matching Algorithm}

The efficacy of our pairing algorithm is demonstrated by the task success rates. As shown in Table~\ref{tab:sim_results_matching}, models trained using our algorithm on both \textit{Perfect} and \textit{Noisy} initial datasets achieved high success rates across all interaction modalities. In contrast, models trained with randomly paired demonstrations exhibited near-total failure. This result shows that a correspondence between demonstrations is a fundamental prerequisite for our method. 

\begin{figure}
    \centering
    \includegraphics[width=0.85\columnwidth]{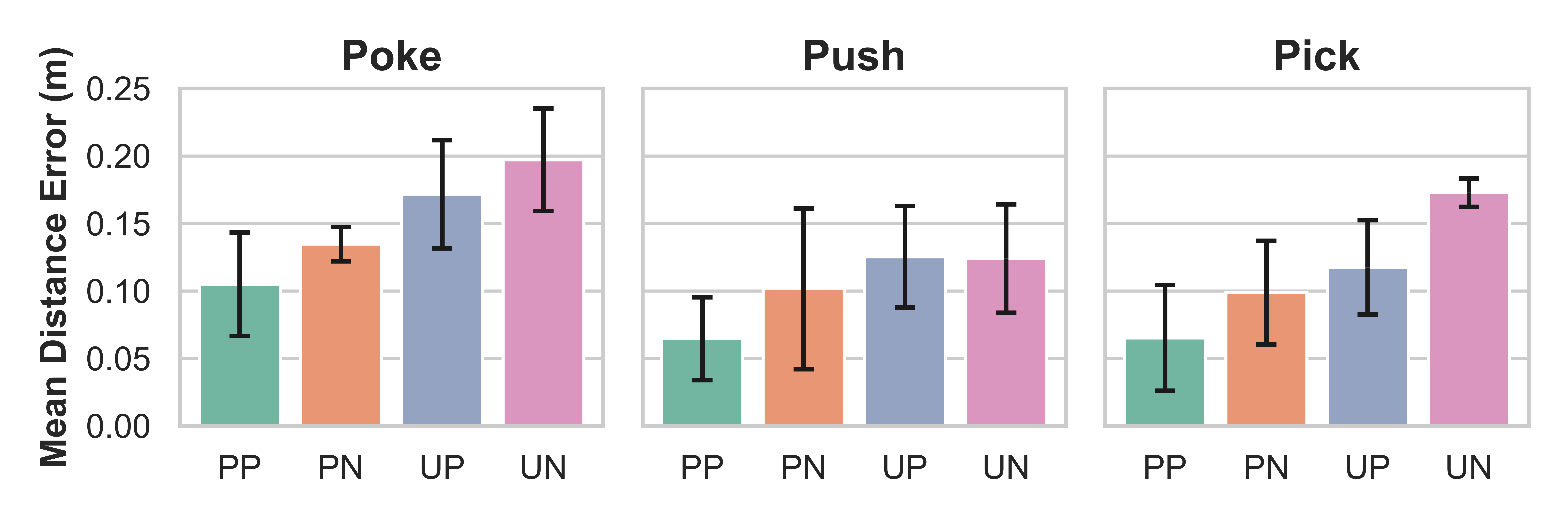}
    \caption{\serdaradd{Final object position error (mean $\pm$ std) for the simulation experiment, across the three tasks and four training conditions}} 
    \label{fig:sim_results}
\end{figure}

Trajectory and object errors are reported in Table~\ref{tab:sim_results_matching} and Fig.~\ref{fig:sim_results}, respectively. For the pick modality, the most complex skill in our setup due to precise gripper-pose alignment, the model trained on \textit{Perfect} dataset yielded significantly lower trajectory error ($p<0.001$) and more accurate final object placement ($p<0.05$), suggesting that high-precision skills are more sensitive to demonstration-pair quality. For the simpler poke action, the \textit{Noisy} dataset significantly reduced trajectory error ($p<0.05$), with no significant final object-position difference (\serdaradd{$p=0.13$}). For push, Perfect and Noisy conditions showed no significant difference in trajectory (\serdaradd{$p=0.06$}) or object-position error (\serdaradd{$p=0.17$}). Overall, perfect demonstrations help complex tasks, while our algorithm still learns robust policies for simpler interactions from imperfect data, comparable to the perfect case. 

\serdaradd{Furthermore, the accuracy of our model under different noise conditions demonstrates the robustness of our pairing algorithm, as shown in Table~\ref{tab:sim_obs_noise}. The matching accuracy decreases with increasing observation noise (highest 100\%, lowest 66\%), remaining near-perfect when $\sigma_\text{cam} \le 0.75 \mathrm{cm}$ and $\sigma_\text{pos} \le 0.5 \mathrm{cm}$. However, we note that the trajectory error increases by $\approx16\%$ on average when matching accuracy stays above 90\%, and by $\approx 67\%$ when it falls below, indicating that mismatched pairs hurt the model's performance.}

\begin{table}[tb]
\centering
\caption{\serdaradd{Inverse-Task Performance over 8 runs in Simulation: Success Rates (Mean) out of 10 trials and RMSE from Ground Truth in degrees (Mean $\pm$ Std).}}
\label{tab:sim_results_matching}
\setlength{\tabcolsep}{4pt}
\begin{tabular}{l ccc ccc}
\toprule
& \multicolumn{3}{c}{Success Rate } & \multicolumn{3}{c}{RMSE (deg)} \\
\cmidrule(lr){2-4} \cmidrule(lr){5-7}
& Poke & Push & Pick & Poke & Push & Pick \\
\midrule
PP   & \textbf{10.0} & \textbf{8.1} & \textbf{8.2} & 3.21$\pm$0.89 & \textbf{8.23}$\pm$\textbf{1.18} & \textbf{3.17}$\pm$\textbf{0.60} \\
PN  & \textbf{10.0} & 6.4& 6.2 & \textbf{2.99}$\pm$\textbf{0.88} & 10.25$\pm$2.15 & 4.36$\pm$1.37 \\
UP & 0.0 & 0.0 & 3.5 & 34.18$\pm$5.06 & 31.63$\pm$4.51 & 9.41$\pm$4.16 \\
UN & 0.9 & 2.5 & 0.2 & 30.58$\pm$8.57 & 22.38$\pm$7.11 & 23.45$\pm$7.79 \\
\bottomrule
\end{tabular}
\end{table}

\begin{table}[t]
\centering
\caption{\serdaradd{Trajectory error (number of incorrect matches out of 60) for various observation noises (in cm), over 3 seeds, 30 cases each.}}
\label{tab:sim_obs_noise}
\begin{tabular}{cccc}
\toprule
\diagbox[width=1.5cm]{$\sigma_{\text{cam}}$}{$\sigma_{\text{pos}}$} & \textbf{0.25} & \textbf{0.50} & \textbf{1.00} \\
\midrule
\textbf{0.50} & 4.63 $\pm$ 0.36 (0) & 4.63 $\pm$ 0.36 (0) & 5.30 $\pm$ 0.30 (4) \\
\textbf{0.75} & 5.52 $\pm$ 0.40 (2)  & 5.37 $\pm$ 0.36 (2)  & 6.30 $\pm$ 1.87 (10) \\
\textbf{1.00} & 5.64 $\pm$ 0.38 (10)  & 6.92 $\pm$ 1.71 (12)  & 12.1 $\pm$ 3.83 (20) \\
\bottomrule
\end{tabular}
\end{table}


\paragraph{Comparison with Baselines}

Trajectory and object errors and success rates for our method and the baselines are shown in Table~\ref{tab:sim_results}. \serdaradd{For \textbf{MVAE-Mask}, we evaluated three forward-observation conditions: ten uniformly sampled forward observations, the first and final observations, and without forward observations (task parameter only). We obtained mean trajectory RMSEs of $7.22$, $7.88$, and $11.48$ degrees, respectively, and reported the lowest-error case. \textbf{MVAE} was conditioned on the full forward observation and the task parameter. For our method, we observed that the trajectory error was essentially invariant to forward conditioning (maximum absolute difference of $10^{-3}$ degrees), and we reported the case with the initial forward observation and the task parameter.} 
\serdaradd{Overall, our method significantly outperformed the best-performing baseline by overall task success rate, DP-Mode, across all three metrics ($p<0.001$ each), despite having $70\times$ fewer trainable weights. Both MVAE baselines attained lower mean trajectory RMSE than \textbf{DP-Mode}, yet their success collapsed on the pick task ($0.8$ and $1.0$ out of $10$), which required precise gripper-pose regression. Also, our method achieved significantly lower trajectory and object errors than both MVAE variants ($p<0.05$ each). In sum, the results imply that the baselines had difficulty learning a generalizable representation of task parameters of novel objects when paired and auxiliary task parameters were available.}

\begin{table}[bt]
\centering
\caption{\serdaradd{RMSE from ground truth trajectory and distance error (DE) for final object position (Mean $\pm$ Std), and Success Rates out of 10 trials (Mean) over 8 runs}}
\label{tab:sim_results}
\begin{tabular}{l cc ccc}
\toprule
& \textbf{Traj. Error} & \textbf{Obj. Error}
& \multicolumn{3}{c}{\textbf{Success Rate}} \\
\cmidrule(lr){2-3}  \cmidrule(lr){4-6}
\textbf{Method} 
  & \textbf{RMSE (deg)} & \textbf{DE (cm)}
  & \textbf{Poke} & \textbf{Push} & \textbf{Pick} \\
\midrule
Ours (UP)  & 25.13 $\pm$ 2.32 & 13.77 $\pm$ 1.87 
           & 0.0           & 0.0           & 3.5           \\
DP-Dual    & 24.68 $\pm$ 1.52 & 16.35 $\pm$ 0.85 
           & ---           & ---           & ---    \\

DP-Mode    & 8.67 $\pm$ 1.08 & 12.50 $\pm$ 2.58 
           & 7.4 & 4.5  & 3.9         \\

MVAE-Mask    & 7.22 $\pm$ 4.83 & 14.55 $\pm$ 3.66 & 6.2 & 6.0  & 0.8        \\

MVAE    & 5.39 $\pm$ 0.83 & 13.11 $\pm$ 1.99 & 7.9  & 6.6  & 1.0 \\

Ours (PP)  & \textbf{4.79 $\pm$ 0.28} & \textbf{7.79 $\pm$ 2.30} 
           & \textbf{10.0}  & \textbf{8.1}  & \textbf{8.2} \\
\bottomrule
\end{tabular}
\end{table}


\paragraph{Extrapolation Capabilities}
A key result of this experiment is the model's ability to extrapolate to objects from the object classes presented in the \finaldiff{Auxiliary dataset. The model's success on} the novel objects was enabled by the auxiliary dataset ($\mathcal{D}_{aux}$), which provided examples of forward task executions for spheres and boxes. By leveraging auxiliary demonstrations, the model integrated new object representations into its latent space, successfully inferring inverse task behaviors such as poking a sphere. The pick modality extrapolation is particularly noteworthy: unlike poke and push, which rely on categorical object distinctions, picking requires regressing the specific orientation of novel boxes (Fig.~\ref{fig:sim_tasks}, bottom row). This successful inference of continuous geometric parameters provides compelling evidence of a rich, structured latent representation of the new object classes.

%% file: algorithm_tikz.tex
\begin{figure}[tb]
    \centering
    \resizebox{0.8\columnwidth}{!}{
    \begin{tikzpicture}[
        node distance=2.0cm and 2.5cm, 
        font=\Large,
        thick,
        >={Stealth[length=2mm]},
        start/.style={draw, rounded corners, fill=gray!20, align=center, minimum height=2.5em, inner sep=8pt},
        decision/.style={draw, diamond, aspect=1.5, fill=blue!5, align=center, inner sep=2pt, font=\large},
        action/.style={draw, rectangle, fill=green!10, rounded corners, align=center, minimum height=2.5em, minimum width=2.2cm, font=\Large},
        line/.style={->, draw, thick, rounded corners=4pt},
        label_edge/.style={midway, fill=white, inner sep=2pt, font=\Large, text=black}
    ]
    
        
        \node[start] (start) {Input \\ Task Parameter};
        \node[decision, right=1.2cm of start] (obj_type) {Object\\Type?};
        
        \node[action, right=6.0cm of obj_type] (pick) {PICK};
        \node[action, above=1.05cm of pick] (poke) {POKE};
        \node[action, below=1.05cm of pick] (push) {PUSH};
        
        \node[decision] (check_horiz) at ($(obj_type)!0.5!(poke |- obj_type) + (0.8, 1.0)$) {Size $> T$?};
        
        \node[decision] (check_cube) at ($(obj_type)!0.5!(push |- obj_type) - (-0.8, 1.0)$) {Size $> T$?};
    
    
        \draw[line] (start) -- (obj_type);
    
        \draw[line] (obj_type.north) -- ++(0, 1.8) -| node[label_edge, pos=0.15] {Sphere} (poke.north);
        
        \draw[line] (obj_type.south) -- ++(0, -1.8) -| node[label_edge, pos=0.15] {Vert. Cyl.} (push.south);
    
        \draw[line] (obj_type.east) -- ++(0.5, 0) |- node[label_edge, pos=0.5] {Horiz. Cyl.} (check_horiz.west);
        
        \draw[line] (obj_type.east) -- ++(0.5, 0) |- node[label_edge, pos=0.5] {Box} (check_cube.west);
    
        \draw[line] (check_horiz.north) |- node[label_edge, pos=0.7] {Yes} (poke.west);
        \draw[line] (check_horiz.east) -| node[label_edge, pos=0.3] {No} (pick.north);
    
        \draw[line] (check_cube.south) |- node[label_edge, pos=0.7] {Yes} (push.west);
        \draw[line] (check_cube.east) -| node[label_edge, pos=0.3] {No} (pick.south);
    
    \end{tikzpicture}
    }
    \caption{\finaldiff{Task controller logic, showing the underlying decision process.}}
    \label{fig:controller_logic}
\end{figure}


%% file: real_world_experiments.tex
\subsection{\serdaradd{Extrapolation in Real Robot}}

\serdaradd{We designed two real-world tasks to test our method. The first evaluates extrapolation across tools and data efficiency by comparing models trained with a full auxiliary set versus a minimal two-demonstration set. The second is a more complex task requiring generalization to different glasses that must be picked up from a table or a dish rack and placed back on the table or into the dish rack. We analyzed how our method's performance scales as the number of extrapolated objects increases and compared its performance with MVAE. We used a 7-DoF \textit{xArm 7} with a standard gripper and an overhead \textit{Intel RealSense} camera.}

\begin{figure}
    \centering
    \includegraphics[width=0.85\columnwidth]{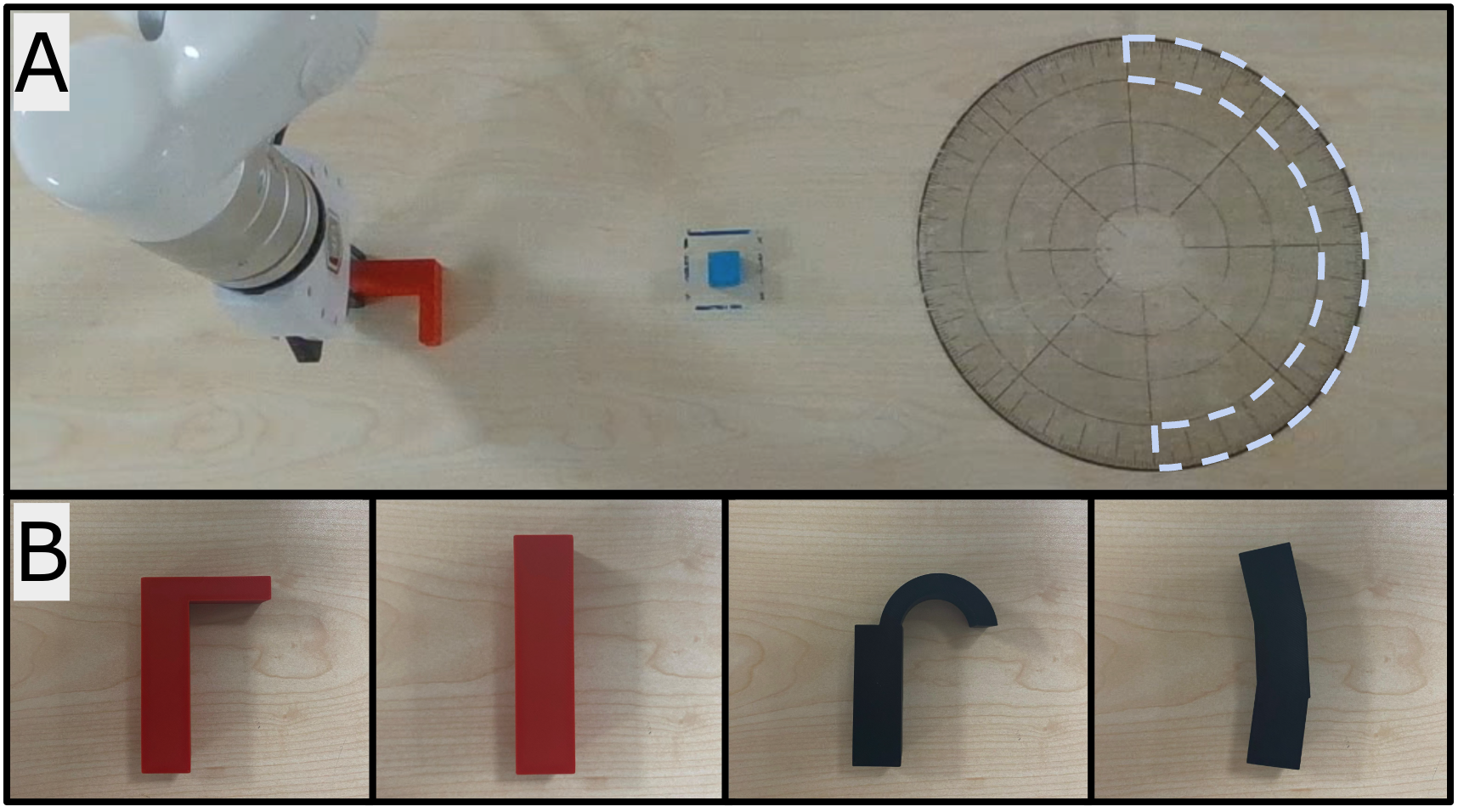}
    \caption{\serdaradd{Setup for the tool experiment.} (A) The tabletop environment, showing the xArm 7 robot, the cube's initial position for the forward task (final position for the inverse task), and the highlighted area on the dial (possible final positions for the forward task). (B) The 3D-printed tools used: L-stick and Stick (red tools), and the novel Hook and Tilted-stick (black tool).}
    \label{fig:real_setup}
\end{figure}

\subsubsection{Tool Extrapolation}

In \textbf{forward task}, the robot uses a tool held in its gripper to push a blue cube from a fixed starting position to a target (x, y) coordinate within a designated area (highlighted in Fig.~\ref{fig:real_setup}.A) on a circular dial. In \textbf{inverse task}, the robot uses the same tool to pull the cube from its position on the dial back to the initial starting position. Four different 3D-printed tools, shown in Fig.~\ref{fig:real_setup}, are used for the task. For each demonstration, we recorded the joint angles of the robot as the SM trajectory, $\tau$. The task parameter, $\psi$, was a vector consisting of the $(x, y)$ coordinates of the blue cube and a feature vector derived from an RGB image of the tool. The coordinates are computed from the workspace image pixel values, and the tool images were preprocessed by segmenting the tool and extracting its edge contours.

\paragraph{Datasets} The \textbf{Initial Sets} ($\mathcal{D}_F, \mathcal{D}_I$) consisting of 40 demonstration pairs, including 20 with Stick and 20 with L-stick, and \textbf{Auxiliary Set} ($\mathcal{D}_{aux}$) consisting of 20 Push demonstrations only, including 10 with a Tilted-stick and 10 with a Hook is demonstrated in the workspace. The state variables ($S_{init}$ and $S_{final}$) were defined as the position of the object and the used tool. The demonstrations in the initial dataset are then paired according to the final position of the object after forward demonstrations, and the initial position of the object before inverse demonstrations. 

\paragraph{Model} The models use an MLP encoder and decoder, and a CNN for the task parameter. We trained the model under two conditions: training with the full Auxiliary set and with a minimal set (2 demonstrations, including 1 with a Tilted-stick and 1 with a Hook). The models (1.7M trainable weights) are trained with $p_{\text{aux}} = 0.25$, batch size 16 for 100K steps using AdamW optimizer (lr: $1\cdot10^{-4}$, weight decay: $1 \cdot 10^{-3}$).

\paragraph{Results}

To evaluate the model, we placed the cube at 10 different points on the circular dial and generated pull trajectories for each novel tool. We also demonstrated 20 trajectories for the inverse task with the Tilted-stick and Hook as ground-truth trajectories for further evaluation. We considered an inverse demonstration successful if the robot manages to move the object from its position on the circular dial to the area surrounded by blue lines (Fig. \ref{fig:real_setup}.A). We used the model trained with the minimal Auxiliary set to generate trajectories. The robot successfully completed the pull task in \textbf{7 out of 10} trials for both novel tools, the tilted stick and the hook, showing that our approach can effectively learn a shared latent space and generalize to novel \finaldiff{task parameters. We also computed} the RMSE between generated and ground-truth trajectories for both full auxiliary and minimal auxiliary (20 and 2 demonstrations, respectively) training conditions. The minimal auxiliary set matches the full set in the trajectory RMSE with no significant difference (full: $8.80^\circ$, min.: $8.62^\circ$, $p=0.72$), highlighting the data efficiency of our method. Furthermore, we analyzed the activations from the CNN used for tool image embedding. The novel Tilted-stick produced a similar embedding to the Stick tool (full: $0.102$ vs. $0.484$; min.: $0.0001$ vs. $-0.340$), while the Hook tool is similar to the L-stick tool (full: $-0.057$ vs. $-1.010$; min.: $2.202$ vs. $0.519$), suggesting that the network learns a semantically meaningful representation of tool geometry.

\begin{figure}
    \centering
    \includegraphics[width=0.85\columnwidth]{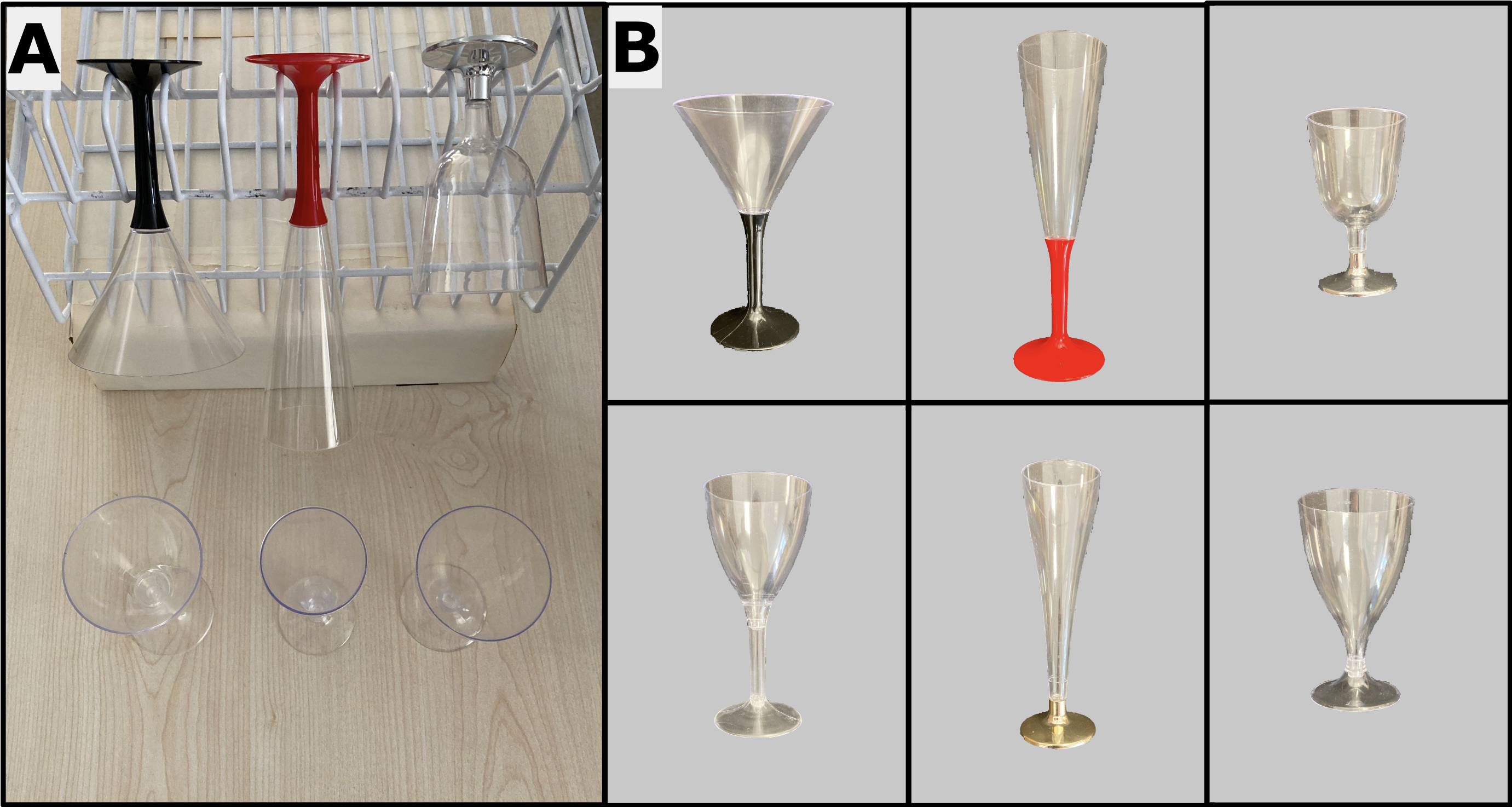}
    \caption{\serdaradd{Setup for the glass hanging experiment. (A) The tabletop environment has a dish rack with nine possible slots. Glasses are hung in the 2nd, 5th, and 8th slots from right to left. (B) The glasses used in the demonstrations.}}
    \label{fig:real_glass_setup}
\end{figure}

\subsubsection{\serdaradd{Object Extrapolation for Glass Hanging}}


\serdaradd{In the \textbf{forward task}, the robot picks one of six glasses from the table and places it inverted into a target dish-rack slot (Fig.~\ref{fig:real_glass_setup}). In the \textbf{inverse}, it removes the same glass and returns it upright to its start position. The task parameter $\psi$ consists of the target-slot scalar and an RGB-image-derived glass feature vector.}

\paragraph{\serdaradd{Datasets}}

\serdaradd{The demonstration set contains 54 pairs: 9 demonstrations for each of six glasses. The three upper-row glasses in Fig.~\ref{fig:real_glass_setup}.B appear in every dataset with full forward–inverse pairs. The three lower-row glasses are extrapolation objects: in each dataset, some of them are held out, meaning only a single forward demonstration, and no inverse counterparts, is provided for those glasses. We hold out combinations of the three, giving C(3,1) + C(3,2) + C(3,3) = 7 datasets with one, two, or three held-out glasses, to study how performance scales with the number of extrapolated objects.}

\paragraph{\serdaradd{Models}}

\serdaradd{Our model uses the same architecture as in the tool experiment, with 1M weights. As a baseline, we train MVAE with 1M weights and the same hyperparameters from the simulation experiments. We use MVAE as the only baseline, since it achieved better or competitive performance compared to the other baselines while requiring significantly shorter training time. For both methods, a separate model is trained for each dataset.}

\begin{table}[t]
\centering
\caption{\serdaradd{Task success under different numbers of extrapolated objects, reported as successful/total trials (success rate)}}
\label{tab:glass_results}
\begin{tabular}{@{}lcccc@{}}
\toprule
\textbf{Model} & \textbf{1 obj.} & \textbf{2 objs.} & \textbf{3 objs.} & \textbf{Total} \\
\midrule
Ours & \textbf{26/27 (96.3)} & \textbf{44/54 (81.5)} & \textbf{20/27 (74.1)} & \textbf{90/108 (83.3)} \\
MVAE & 0/27 (0.0) & 6/54 (11.1) & 1/27 (3.7) & 7/108 (6.5) \\
\bottomrule
\end{tabular}
\end{table}

\paragraph{\serdaradd{Results}}

\serdaradd{A trial is counted as successful if the robot retrieves the glass from the target rack slot and returns it. For each held-out glass and model, we generated 9 inverse trajectories; our method achieved $83.3\%$ success, versus $6.5\%$ for MVAE (Table~\ref{tab:glass_results}). When measured against human demonstrations, both methods produced comparable trajectory RMSE (ours: $9.57^\circ$, MVAE: $9.99^\circ$; $p=0.72$). However, they differed substantially in trajectory jerk RMSE (ours: $0.088$ $\mathrm{^\circ/s}^3$, MVAE: $0.401$ $\mathrm{^\circ/s}^3$; $p<0.001$), indicating that the MVAE's trajectory predictions were accurate but not smooth enough to manipulate the glasses reliably. A common failure mode for MVAE was visible jerking that dislodged the glass while picking it from the rack slot. Notably, our method performed nearly perfectly (96.3\%) in the single held-out case. However, the success rate monotonically decreased in two- and three-held-out cases ($81.5\%$, $74.1\%$), as increasing the number of extrapolated objects reduced the geometric diversity of the remaining glasses in paired demonstrations.}

%% file: limitations.tex



%% file: conclusion.tex
\section{Conclusion}

We proposed a novel joint learning framework that enables generalization to novel task parameters by transferring knowledge between a jointly learned forward-inverse task pair. Through comprehensive experiments on synthetic, simulated, and real-world data, we showed that our demonstration-pairing algorithm is critical for establishing a coherent common latent space\serdaradd{, and robust to observation noise. We also showed that our architectural extensions and training methodology enable a robot to efficiently transfer knowledge and extrapolate manipulation skills to novel task parameters, such as objects and tools, outperforming diffusion-based and MVAE baselines.} Our approach provides a data-efficient solution within imitation learning by leveraging the assumption that novel task parameters can be projected to known behaviors. A limitation of our framework is that novel auxiliary task parameters can be mapped to behaviors already present in the demonstrated SM space of the forward-inverse task pair, excluding scenarios where inversion requires unobserved sub-actions. \serdaradd{In future work, we plan to extend the evaluation of our method under more open-ended scenarios, such as complex assembly-disassembly tasks.
} Although the presented method is confined to forward-inverse tasks where an intuitive state-based pairing algorithm is effective, the core principle can be extended to any related task pair. This would likely require a more complex pairing algorithm capable of discovering the underlying structural relationship between tasks. 
